%% file: main.tex
\documentclass[conference]{IEEEtran}
\IEEEoverridecommandlockouts
\usepackage{cite}
\usepackage{amsmath,amssymb,amsfonts}
\usepackage{algorithmic}
\usepackage{algorithm}
\usepackage{graphicx}
\usepackage{verbatim}
\usepackage{textcomp}
\usepackage{xcolor}
\usepackage{url}
\usepackage{bm}
\usepackage{subfigure}
\usepackage{flushend}
\usepackage{placeins}
\usepackage{capt-of}
\renewcommand{\algorithmicrequire}{\textbf{Initialize:}}

\def\BibTeX{{\rm B\kern-.05em{\sc i\kern-.025em b}\kern-.08em
    T\kern-.1667em\lower.7ex\hbox{E}\kern-.125emX}}
\begin{document}

\title{(DEMO) Deep Reinforcement Learning Based Resource Allocation in Distributed IoT Systems
}
\author{Aohan Li\IEEEauthorrefmark{1} and Miyu Tsuzuki\IEEEauthorrefmark{1}
\\
\IEEEauthorrefmark{1}Graduate School of Informatics and Engineering, The University of Electro-Communications, Tokyo, Japan\\

aohanli@ieee.org, t2531097@edu.cc.uec.ac.jp

}

\maketitle

\begin{abstract}
Deep Reinforcement Learning (DRL) has emerged as an efficient approach to resource allocation due to its strong capability in handling complex decision-making tasks.
However, only limited research has explored the training of DRL models with real-world data in practical, distributed Internet of Things (IoT) systems. 
To bridge this gap, this paper proposes a novel framework for training DRL models in real-world distributed IoT environments. 
In the proposed framework, IoT devices select communication channels using a DRL-based method, while the DRL model is trained with feedback information—specifically, Acknowledgment (ACK) information—obtained from actual data transmissions over the selected channels. 
Implementation and performance evaluation, in terms of Frame Success Rate (FSR), are carried out, demonstrating both the feasibility and the effectiveness of the proposed framework.


\end{abstract}


\input{1_Introduction}
\input{3_ProPosed}
\input{4_Results}

\input{5_Conclusions}

\section*{Acknowledgment}
This work was supported by JSPS KAKENHI Grant Number 25K17606.

\input{references}
\end{document}

%% file: 1_Introduction.tex
\section{Introduction}
\label{sect:introduction}

In recent years, the number of Internet of Things (IoT) devices has grown rapidly, driven by advancements in communication technologies such as LoRa, Sigfox, and NB-IoT, the declining cost of sensors and embedded systems, and the application of artificial intelligence in data processing.
The widespread adoption of IoT across various domains, including smart cities, industrial automation, smart homes, and healthcare, has created significant opportunities for the development of efficient and scalable IoT systems, while also introducing challenges such as spectrum congestion \cite{A1}.

Effective resource allocation (RA) methods are essential to ensure high-quality communication for a large number of IoT devices.
Compared to centralized RA methods, distributed approaches can enhance spectrum and energy efficiency, as they do not require the transmission of relevant information to a central processor. 
Deep Reinforcement Learning (DRL) has proven to be an effective method for distributed RA, owing to its ability to tackle complex decision-making problems \cite{A2}.
However, most deep learning-based methods are challenging to implement in practice, as they rely on prior state information.
To solve this problem, feasible methods based on Acknowledgment (ACK) information for IoT systems are proposed in \cite{A3,A4}. However, these studies do not address how to implement these methods in practice, nor do they consider the impact of real-world environments on their performance.

To bridge the gap discussed above, this paper presents a DRL-based RA framework, builds a practical system, and preliminarily validates its effectiveness.

%% file: 3_ProPosed.tex
\section{Proposed Framework}

\label{sec:sec3}

\begin{figure}
\centering
\includegraphics[width=8cm]{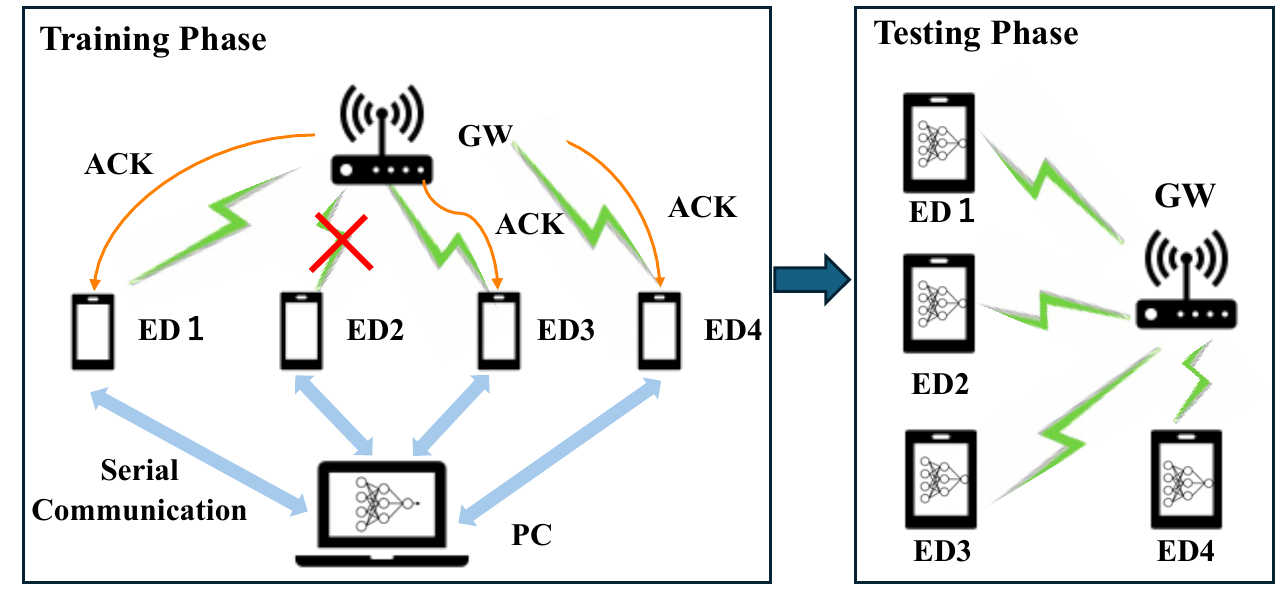}
\caption{Proposed Framework}
\label{fig:s1}
\end{figure}

This section presents the proposed DRL-based RA framework.
As illustrated in Fig. \ref{fig:s1}, the framework consists of two phases: training and testing. 
In the training phase, the IoT end devices (EDs) train their DRL models with the assistance of a Personal Computer (PC).
Specifically, the DRL model for each ED is executed on the PC. 
Before each transmission, the PC performs RA, i.e., channel selection, for each ED based on its DRL model.
The selected channel is then transmitted to the ED
via serial communication. 
The ED uses the assigned channel to transmit data to the Gateway (GW).
If the transmission is successful, the ED receives an ACK information; otherwise, it obtains feedback information regarding the transmission failure.
This feedback is subsequently sent to the PC to update the DRL model.
The above process is repeated until $N$ training iterations are completed.
In the testing phase, the trained DRL model of each ED is employed directly for channel selection during data transmission.

In the training phase of our proposed framework, the components of DRL are defined as follows.
The agent is the IoT ED, while the state and reward are determined based on feedback information.
Specifically, if an ACK information is received after data transmission using the selected channel, both the next state and the reward are set to 1; otherwise, they are set to 0. 
The action corresponds to channel selection, and the $\epsilon$-greedy method is adopted as the policy.
During training, $\epsilon$ decreases linearly according to $\epsilon_n = 1 - \frac{n-1}{N-1}$, where $n$ denotes the current episode.
To improve learning stability and performance, Double Deep Q-Learning is employed to train the model. 
The overall procedure of the training phase for each IoT ED is summarized in Algorithm 1.

\begin{algorithm}
\small

\caption{Training Phase}
\begin{algorithmic}[1]
\renewcommand{\algorithmicrequire}{\textbf{Initialize:}}
\REQUIRE main network weights $\theta_n=\theta^{\rm init}_n$, target network weights $\theta^{-}_n=\theta_n$, time step $t= 0$, learning episode $n=0$, state $s=0$.

\WHILE{$n \leq N$}
    \WHILE{$t \leq T$}
     \STATE Select action $a \in \mathcal{A}$ with $\epsilon$-greedy policy
            \[
            a = 
            \begin{cases}
                \underset{a \in \mathcal{A}}{\arg \max} Q(s, a; \theta_n), & \text{with prob. }1-\epsilon_n
                \\
                \underset{a\in \mathcal{A}}{\operatorname{random}}(a), &\text{with prob. }\epsilon_n
                \end{cases};
            \]
     \STATE Packet transmission use the selected channel;
    \STATE Observe next state $s'$ , and the corresponding reward $r$ with:
            \[    
           s'= r= 
                \begin{cases}
                1, & \text{if ACK received} \\
                0, & \text{otherwise}
                \end{cases}
            ;
            \label{eq:reward}
            \]
    \STATE Store the transaction $(s, a, r, s')$ in replay buffer $D_n$;
                \STATE Sample mini-batch $B_n$ from $D_n$;
                
                \STATE For each transaction $(s, a, r, s') \in B_n$, compute target value $Q^*(s, a)$:
                \[Q^*(s, a) = r + \gamma Q(s', \max_{a'} Q(s', a'; \theta_n); \theta_n^-);\]
                \STATE Update $\theta_n$ by loss   
                \[L(\theta_n)=\mathbb{E}{[(Q^*(s, a) - Q(s, a; \theta_n) ) ^ 2]}.\]
                \STATE $t = t + 1$  
                \STATE Every $C$ steps sync weights: $\theta_n^- \xleftarrow{} \theta_n$;
\ENDWHILE
  
    \STATE $n = n + 1$; $t=0$

 \STATE Update $\epsilon_n=1-\frac{n-1}{N-1}$
            
\ENDWHILE
\end{algorithmic}
\end{algorithm}

%% file: 4_Results.tex
\section{Implementation}
\label{sect:simulation result}

\begin{figure} [!t]
\centering
\includegraphics[height=50mm,width=60mm]{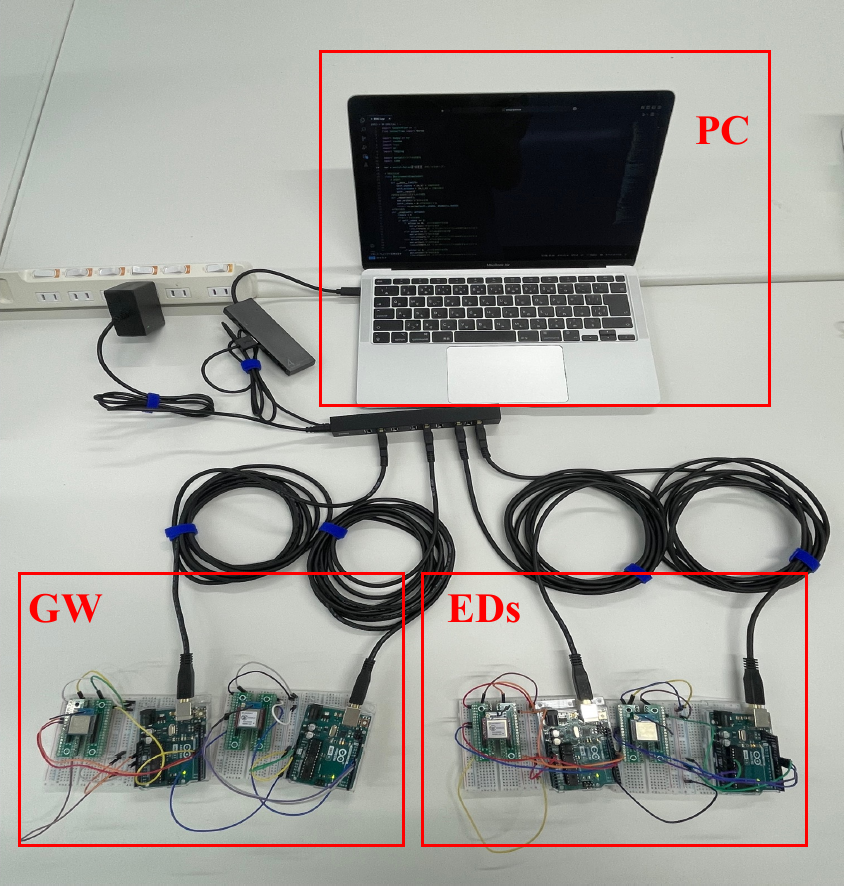}
\caption{Experimental Environment}
\label{fig:implementation}
\end{figure}

Fig. \ref{fig:implementation} illustrates the implementation of the proposed framework, where the number of IoT EDs is set to 2.
The GW can receive data from two channels with frequencies of 922.8 MHz and 923.2 MHz. The IoT EDs select a channel from three available channels with frequencies of 922.4 MHz, 922.8 MHz, and 923.2 MHz, each having a bandwidth of 125 kHz.
In the implementation, both the IoT EDs and the GW are equipped with an Arduino Uno controller and a Long Range (LoRa) module, ES920LR, with a spreading factor set to 7. 
The learning rate $r$ and the discount factor $\gamma$ are set to 0.01 and 0.6, respectively. The number of episodes $N$ is set to 500, and the number of transmissions per episode $T$ is set to 20.
The mini-batch $B_n$ and weights update frequency $C$ are set to 16 and 10, respectively.

\begin{center}
\includegraphics[width=70mm]{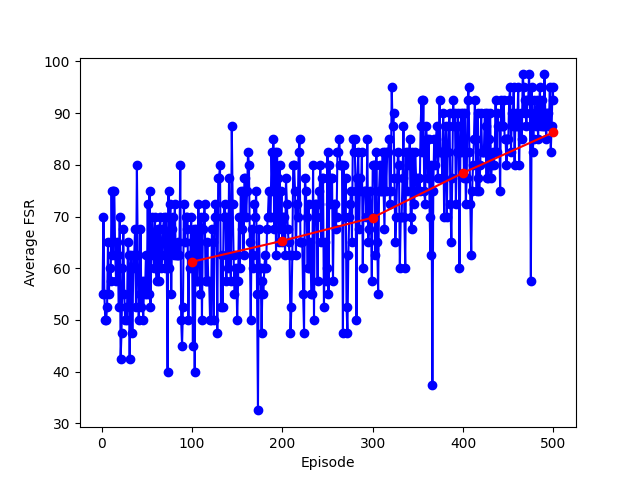}
    \captionof{figure}{Average FSR during the training phase}
    \label{fig:FSR}
\end{center}

Fig. \ref{fig:FSR} presents the average FSR of two IoT EDs during the training phase for each episode. 
The red points in the figure represent the average FSR for every 100 episodes.
As shown in Fig. \ref{fig:FSR}, the average FSR of the IoT EDs gradually increases as training progresses, demonstrating the effectiveness of the training phase in the proposed framework. 
This trend occurs because the DRL method can be effectively trained using environmental feedback.
Furthermore, the trained model is evaluated in the testing phase. The results indicate that the average FSR reaches 91\% over 100 episodes, which represents a 17\% improvement compared with the $\epsilon$-greedy method without a training phase.

%% file: 5_Conclusions.tex
\section{Conclusion}
\label{sect:conclusion}
This paper presents a framework demonstrating the feasibility and effectiveness of implementing DRL-based RA. In future work, we aim to accelerate the convergence of the training phase, enhance the stability of the system, and extend this approach to high-density, scalable, and distributed IoT networks.